\documentclass[11pt]{article}
\usepackage{acl2015}
\usepackage{times}
\usepackage{url}
\usepackage{latexsym}
\usepackage{hyperref}

\usepackage{times}
\usepackage{latexsym}

\usepackage[T1]{fontenc}

\usepackage[utf8]{inputenc}
\usepackage{comment}
\usepackage{microtype}
\usepackage{listings}
\usepackage{amssymb}
\usepackage{hyperref}
\newcommand{\mytt}[1]{\texttt{\detokenize{#1}}}
\newcommand{\myit}[1]{\textit{\detokenize{#1}}}

\title{CLaCLab at SocialDisNER: Using Medical Gazetteers for Named-Entity Recognition of Disease Mentions in Spanish Tweets}

\author{Harsh Verma, Parsa Bagherzadeh, Sabine Bergler \\
        CLaC Labs, Concordia University \\
        \{h\_ver, p\_bagher, bergler\} @cse.concordia.ca}

\date{}

\begin{document}
\newcounter{ex}
\newenvironment{expl}{ \begin{enumerate} \refstepcounter{ex} \item[(\theex)] \begin{it}} {\end{it} \end{enumerate} }

\maketitle
\begin{abstract}
 This paper summarizes the CLaC submission for SMM4H 2022 Task 10 which concerns the recognition of diseases mentioned in Spanish tweets. Before classifying each token, we encode each token with a transformer encoder using features from Multilingual RoBERTa Large, UMLS gazetteer, and DISTEMIST gazetteer, among others. We obtain a strict F1 score of 0.869, with competition mean of 0.675, standard deviation of 0.245, and median of 0.761. Our code is available \href{https://github.com/harshshredding/SMM4H-2022-Social-Dis-NER-Submission}{here}.
\end{abstract}

\section{Motivation}

Finding mentions of diseases in tweets in all languages has become an important tool for epidemiologists, especially in times of a pandemic. SocialDisNER \cite{gasco2022socialdisner} at SMM4H 2022 concerns the recognition of disease mentions in Spanish tweets. A disease mention can include both lay and professional language and may be located in hashtags or usernames as well as the tweet text. Disease entities are underlined in Example~\ref{data}:

\begin{expl} \label{data}
Pasos sábado 23 de mayo! ... Sumemos 1 millón de pasos por la \underline{epilepsia}, \#1MillonDePasos, \#\underline{Epilepsia}, \#ADosMetrosDeDistancia  \#Investiga\underline{Epilepsia} @Reto\underline{Dravet} \#juntosesmejor
@Fundacion\underline{Dravet}
\end{expl}

\section{System}

Our contribution consists of a simple pipeline incorporating four main components: a standard tokenizer for splitting hashtags and username tokens with an ad hoc extension for disease recognition, word embeddings for Spanish and English from RoBERTa Large, four gazetteer lists extracted from relevant domain resources and represented as one hot vectors, and a linear classifier that produces BIO (\emph{beginning, inside, outside}) tags for each token for recognition of multi-token disease mentions. The simplicity of this pipeline and its knowledge injection from readily available domain resources rather than training purely from training data make our system's strength.


\paragraph{Tokenization}

Sequence analysis classifying each token into BIO tags requires good tokenization. We pre-process the data using ANNIE \cite{Cunningham2002} and the Twitter English Tokenizer \cite{twitie}. 

Hashtags and usernames  usually include several words strung together without separating characters, e.g.  \#Investiga\underline{Epilepsia}. Since disease mentions can occur in these word composites, their individual word components have to be separated. Camel casing makes splitting this example into \textit{Investiga} and \textit{Epilepsia} easy, but for hashtags like \#nosolohay\underline{covid} only knowledge of words in Spanish leads to the desired split into \textit{no}, \textit{solo}, \textit{hay} and \textit{covid}. Because disease names are not always part of the general vocabulary of a language, disease gazetteers are a lightweight means to inject such domain knowledge into a general machine learning system.

A gazetteer lists entities of a particular type. For example, a disease gazetteer would contain a list of disease names like COVID, Hepatitis, Epilepsia etc. Ad hoc gazetteer lists that are written for a specific task often suffer from limited coverage. We circumvent our own limitations by using four gazetteer lists compiled from extant resources:

\begin{description}
    \item[\textit{GoldGaz}:] disease names compiled from the  gold annotations of SocialDisNER training and validation data
    
    \item[\textit{DistemistGaz}:] \cite{miranda2022overview}   Spanish disease gazetteer  compiled from Snomed-CT \cite{snomed}
    
    \item[\textit{SilverGaz}:] disease gazetteer compiled from silver standard data made available for SocialDisNER \cite{gasco2022socialdisner}
    
    \item[\textit{UmlsGaz}:]  disease gazetteer of Spanish and English disease terms we compiled from UMLS \cite{UMLS}. Semantic type T047 represents \emph{Disease or Syndrome}, identifying all UMLS concepts representing diseases.
\end{description}

\paragraph{Twitter tokenizer extension} The GATE Twitter English Tokenizer \cite{twitie} identifies hashtags and usernames, but doesn't split them. To split usernames and hashtags, we identify the longest substring in a hashtag/username that matches an entity in our gazetteer,  then treat that substring as a separate token. The extension matches first against GoldGaz, then DistemistGaz, followed by UmlsGaz, and finally SilverGaz.

\paragraph{Classification}
We frame the named entity recognition (NER) task as a sequence labeling task with the BIO (\emph{beginning, inside, outside}) tagging scheme. In other words, we classify each token as either \mytt{B} (token is the beginning of a disease name), \mytt{I} (token lies inside a disease name) or \mytt{O} (token lies outside a disease name).

\paragraph{Model}
Preprocessing produces $S$, a list of $t$ tokens. $S$ is fed into a pretrained XLM RoBERTa Large \cite{xlm} model, which returns the matrix $\textbf{H} \in \mathbb{R}^{t \times d}$, with $d = 1024$. The $i^{th}$ row of $\textbf{H}$ is a vector of size $d$ and corresponds to the contextualized embedding of the $i^{th}$ token. 

We use the GATE ANNIE Gazetteer plugin to find tokens that match  terms
in UmlsGaz exactly (case-insensitively),  creating $\textbf{G}_{\textit{umls}} \in \mathbb{R}^{t \times 2}$, a matrix where each row is a one hot vector indicating whether the $i^{th}$ token matches with \myit{UmlsGaz} or not. Similarly, we create $\textbf{G}_{\textit{silver}}$ and $\textbf{G}_{\textit{distemist}}$ corresponding to \myit{SilverGaz} and \myit{DistemistGaz} respectively. Then,
$\textbf{H}$ is row-wise concatenated with $\textbf{G}_{\textit{umls}}$, $\textbf{G}_{\textit{distemist}}$, and $\textbf{G}_{\textit{silver}}$ to produce $ \textbf{Z} \in \mathbb{R}^{t \times 1030}$. $\textbf{Z}$ is then fed into a 6-layer Transformer Encoder \cite{vaswani2017attention} with positional encoding and 10 attention heads per layer. The Transformer Encoder outputs $\textbf{Y} \in \mathbb{R}^{t \times 1030}$ where the $i^{th}$ row represents the encoded representation of the $i^{th}$ token. $\textbf{Y}$ is fed into a linear classifier which produces $\textbf{I} \in \mathbb{R}^{t \times 3}$, classifying each token into one of the 3 classes corresponding to the BIO tagging scheme. This represents our submission system $M_{\myit{sub}}$.

\paragraph{Training}
XLM RoBERTa large is fine-tuned on the training data using the Adam optimizer \cite{adam} with a learning rate of \mytt{1e-5} and early stopping. Training for 10 epochs takes 2.5 hours on one Nvidia RTX 3090 gpu.

\section{Ablation and test results}
The evaluation is by strict F1 score, requiring  exactly matching  gold spans without  partial credit. 

Our submission model $M_{\myit{sub}}$ was described above in Section 2.2. Let $M_{\myit{no-gaz}}$ be the same model as $M_{\myit{sub}}$ but without the one-hot features generated from gazetteer matching.
Let $M_{\myit{no-tok}}$ be the same model as $M_{\myit{sub}}$ but without the tokenizer extension using disease gazetteers (hashtags and usernames are tokenized with the GATE Hashtag Tokenizer only). Finally, let $M_{\myit{RoBERTa}}$ be the baseline model without one-hot features, custom tokenization, and transformer encoder -- the output of the language model $\textbf{H}$ is fed directly into the classifier.


\begin{table}[htb]
\begin{tabular}{lccc}
\textbf{Model}                & \textbf{F1 score} & \textbf{Precision} & \textbf{Recall} \\
\hline
$M_{\myit{RoBERTa}}$ & 0.880 & 0.856 & 0.905\\
$M_{\myit{no-gaz}}$ & 0.888 & 0.875 & 0.902\\
$M_{\myit{no-tok}}$ & 0.888 & 0.884 & 0.892\\
$M_{\myit{sub}}$  & 0.892  & 0.882 & 0.900\\

\end{tabular}
\caption{Ablation on development set}
\label{validation_performance}
\end{table}

We see that $M_{\myit{sub}}$ improves on $M_{\myit{RoBERTa}}$ by only 1.2\%. Omitting gazetteers or the tokenizer extension each looses only 0.4\% over $M_{\myit{sub}}$.  $M_{\myit{no-gaz}}$ looses precision while $M_{\myit{no-tok}}$ looses recall against the submission system.

Table~\ref{test_performance} shows competition performance of $M_{\myit{sub}}$ on the test set. 
We see that performance on the test set decreased by only 2.3\%, highlighting the robustness of our technique.

\begin{table}[htb]
\begin{tabular}{lccc}
\textbf{Model}  & \textbf{F1 score} & \textbf{Precision} & \textbf{Recall}\\
\hline
$M_{\myit{sub}}$ & 0.869 & 0.851 & 0.888\\
Mean & 0.675 & 0.680 & 0.677\\
Std. dev. & 0.246 & 0.245 & 0.254\\
Median & 0.761 & 0.758 & 0.780

\end{tabular}
\caption{Competition results on test set}
\label{test_performance}
\end{table}

\section{Conclusion}
For the task of detecting disease mentions, a general large, 
pre-trained language model (XLM RoBERTa large) was enhanced with task-oriented preprocessing (splitting hashtags into component words) and lookup in available quality word lists. This combination was effective and in competition placed our system nearly 20\% above the mean.

\bibliographystyle{acl}
\bibliography{custom}

\end{document}